\documentclass[runningheads]{llncs}
\usepackage{graphicx}

\usepackage[utf8x]{inputenc}

\usepackage{tikz}
\usepackage{comment}
\usepackage{amsmath,amssymb} 
\usepackage{color}
\usepackage{url}
\usepackage{hyperref}
\usepackage[capitalise]{cleveref}
\usepackage{booktabs}       
\usepackage{multirow}

\usepackage[accsupp]{axessibility}  

\begin{document}
\pagestyle{headings}
\mainmatter
\def\ECCVSubNumber{17}  

\title{Hierarchical I3D for Sign Spotting} 

\titlerunning{Hierarchical I3D for Sign Spotting}
%
\author{Ryan Wong \and
Necati Cihan Camg\"{o}z\index{Camg\"{o}z, Necati Cihan} \and
Richard Bowden}
\authorrunning{R. Wong et al.}
%
\institute{University of Surrey\\
\email{\{r.wong,n.camgoz,r.bowden\}@surrey.ac.uk}}
\maketitle

\begin{abstract}

Most of the vision-based sign language research to date has focused on Isolated Sign Language Recognition (ISLR), where the objective is to predict a single sign class given a short video clip. Although there has been significant progress in ISLR, its real-life applications are limited. In this paper, we focus on the challenging task of Sign Spotting instead, where the goal is to simultaneously identify and localise signs in continuous co-articulated sign videos. To address the limitations of current ISLR-based models, we propose a hierarchical sign spotting approach which learns coarse-to-fine spatio-temporal sign features to take advantage of representations at various temporal levels and provide more precise sign localisation. Specifically, we develop Hierarchical Sign I3D model (HS-I3D) which consists of a hierarchical network head that is attached to the existing spatio-temporal I3D model to exploit features at different layers of the network. We evaluate HS-I3D on the ChaLearn 2022 Sign Spotting Challenge - MSSL track and achieve a state-of-the-art $0.607$ F1 score, which was the top-1 winning solution of the competition.

\end{abstract}

\section{Introduction}

Sign Languages are visual languages that incorporate the motion of the hands, facial expression and body movement \cite{braem2001hands}. They are the primary form of communication amongst deaf communities with most countries having their own sign languages with different dialects across different regions. Although there are many commonalities across sign languages in terms of linguistics and grammatical rules, each has a very different lexicon \cite{sandler2006sign}.

There has been increasing interest in computational sign language research. A popular research topic has been Isolated Sign Language Recognition (ISLR), where the goal is to identify which single sign is present in a short isolated sign video clip \cite{joze2018ms,li2020word}. Although there are still challenges to solve, such as signer-independent recognition \cite{sincan2020autsl}, the real-life applications of ISLR are limited. 

In this work we focus on the closely related field of Sign Spotting, where the objective is to identify and localise instances of signs within a co-articulated continuous sign video. Sign spotting is beneficial for several real-life applications, such as Sign Content Retrieval, where spotting models are used to search through large unlabelled corpora to locate instances of signs. 

Current sign spotting approaches can be categorized under two groups.
The first is dictionary based sign spotting approaches where given an isolated sign, the objective is to identify and locate co-articulate instances of that sign in a continuous sign video. This usually involves one-shot / few shot learning where minimal annotated examples are available \cite{varol2022scaling}.

The second group of sign spotting approaches align closer to ISLR, but instead of a sign video containing an isolated sign, the video segment is usually longer with one or multiple instances of co-articulated signs that needs to be identified from a set vocabulary. This involves multiple shot supervised learning where there are multiple examples of a set vocabulary within a larger corpus of continuous sign videos. 

\begin{figure*}
    \centering
    \includegraphics[width=1.0\textwidth]{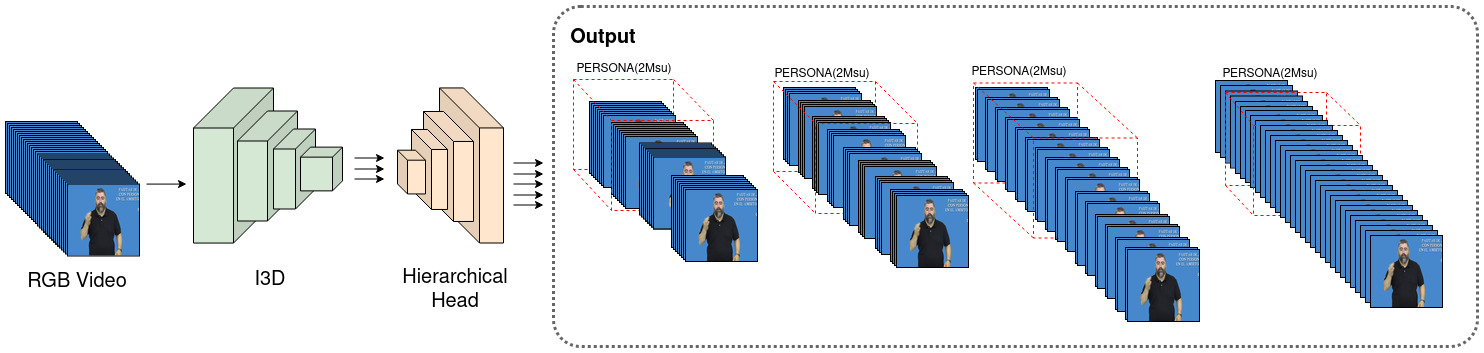}
    \caption{Concept of the Hierarchical Sign I3D model which takes an input video sequence and predicts the localisation of signs at various temporal resolutions}
    \label{fig:main_diag}
\end{figure*}

In this work we build up on the latter group of sign spotting. To address the limitations of the previous approaches, we propose a novel hierarchical spatio-temporal network architecture, named a Hierarchical Sign I3D model (HS-I3D), and identify coarse-to-fine temporal locations of signs in continuous sign videos as shown in \cref{fig:main_diag}. 

HS-I3D comprises of a backbone and a head. Although our approach can be used with any spatio-temporal backbone, we've chosen I3D due to its success in related sign tasks \cite{varol2022scaling}, which enables the use of other pretrained SLR models.
The additional hierarchical spatio-temporal network head has the ability to predict sign labels at the frame level for better estimation of the boundaries between signs.

The main contributions of this work can be summarised as:
\begin{enumerate}
    \item We introduce a novel hierarchical spatio-temporal network head which can be attached to existing spatio-temporal sign models to learn the coarse-to-fine temporal locations of signs.
    \item We demonstrate the importance of incorporating random sampling techniques during training and show the impact and trade-off it has between precision and recall. 
    \item Our architecture achieves state-of-the-art results on the 2022 ChaLearn Sign Spotting Challenge in the multiple shot supervised learning (MSSL) track.
\end{enumerate}

\section{Related work}

\subsection{Sign Language Recognition}

Over the last few decades, significant progress has been made towards Sign Language Recognition. Traditional feature engineered approaches, such as hand shape and motion modeling techniques \cite{bilal2011vision,cooper2011sign,fillbrandt2003extraction,holden2001visual} have been replaced by data driven, machine learning approaches. These data driven approaches require large annotated datasets therefore many ISLR datasets have been created, including but not limited to Turkish Sign Language (TID) \cite{sincan2020autsl}, American Sign Language (ASL) \cite{joze2018ms,li2020word}, Chinese Sign Language \cite{zhang2016chinese} and British Sign Language (BSL) \cite{albanie2020bsl,cormier2012corpus}.

Most of the current ISLR approaches use either raw RGB videos or pose-based input. The pose based input utilizes human pose estimators such as OpenPose \cite{cao2017realtime} and MediaPipe \cite{lugaresi2019mediapipe}, which distill the signer to a set of keypoints. By using keypoints, irrelevant appearance information is discarded, such as the background and a person's visual appearance. Various models have been developed to allow keypoint input, Pose$\rightarrow$Sign \cite{albanie2020bsl} makes use of a 2D ResNet architecture \cite{he2016deep} and found that the keypoint inputs underperformed compared to RGB based approaches. More recently, Graph Convolutional Networks (GCNs) using human keypoints have achieved comparable results to RGB models \cite{jiang2021skeleton}.

RGB based approaches, which use the raw frames as input, have been extended to spatio-temporal architectures, building on existing action recognition models, such as 3DCNNs \cite{qiu2017learning} and more recently the I3D model \cite{carreira2017quo}. Such architectures achieve strong classification performance on ISLR datasets \cite{jiang2021skeleton,joze2018ms,li2020word}.\looseness=-1

Like in other areas of computer vision, transfer learning has been shown to be effective in improving results of sign language recognition \cite{albanie2020bsl}, which is especially important for transferring domain knowledge across different sign languages. Motivated by this, we use a pretrained I3D model as the backbone model for our proposed Hierarchical Sign I3D model which allows us to leverage models pretrained on larger scale ISLR dataset.

\subsection{Sign Spotting}

While ISLR aims to identify an isolated sign in a given sequence, Sign Spotting requires identification of both the start and end of a sign instance from a set vocabulary within a continuous sign video. Early methods utilized hand crafted features, such as thresholding-based approaches using Conditional Random Fields, to distinguish the difference between signs in the vocabulary and non-sign patterns \cite{yang2008sign}. Another technique detected skin-coloured regions in frames and utilized temporal alignment techniques, such as Dynamic Time Warping \cite{viitaniemi2014s}. Sequential Interval Patterns were also proposed, which used hierarchical trees to learn a strong classifier to spot signs \cite{ong2014sign}.

One-shot approaches, which use sign dictionaries have recently been explored, where given a set vocabulary from a dictionary, the objective is to locate these signs within a continuous sign video. Sign-Lookup makes use of a transformer-based network which utilises cross-attention to spot signs from a query dictionary to the target continuous video \cite{jiang2021looking}.

Most ISLR models are trained on datasets that are collected specifically for the task, where the signs are slower and not co-articulated. This makes transfer learning from such models to tasks that focus on co-articulated sign videos, such as sign spotting, difficult. However, there have been recent work \cite{li2020transferring} which focus on transferring knowledge from models trained on isolated sign language recognition datasets to find signs in continuous co-articulated sign videos.

In this work we focus on multiple shot supervised learning using the LSE eSaude UVIGO dataset \cite{Enriquez:ECCVW:2022}, where we have multiple instances of the precise locations of signs from a vocabulary to be spotted in a co-articulated sign video. This allows transfer learning to address the domain adaptation problem from models trained on ISLR datasets to the sign spotting dataset.

\section{Approach}

The HS-I3D model consists of a spatio-temporal backbone based on the Inception I3D model with a hierarchical network head to locate sign instances from a set vocabulary. We first briefly give an overview of the feature extraction method used to extract spatio-temporal representations at different resolutions. Then we introduce the hierarchical network head to make predictions at different temporal resolutions. Finally we describe the learning objectives for the localisation of signs.

\subsection{I3D Feature Extraction Layers}

The I3D model is a general video classification architecture where given a sequence of frames, one class is predicted. This does not naturally allow for the localisation of signs or multiple predictions within a sequence.

To address the objective of localisation, we use the I3D model to extract features at different spatio-temporal resolutions. Instead of taking the output at the final layer, after spatio-temporal global average pooling and applying a fully connected layer for classification, we take the output before global average pooling and additional intermediary features obtained from the outputs before the spatio-temporal max pooling layers in the I3D model. We therefore obtain three spatio-temporal features, each with different temporal resolutions, where features extracted from earlier layers have larger temporal dimensions.

For the rest of this work we assume that the input to the I3D model contains 32 consecutive frames with a resolution of $224\times224$. For a given input sequence, the dimensions of the features extracted from the I3D model are as follows $1024 \times 4 \times 7 \times 7$, $832 \times 8 \times 14 \times 14$ and $480 \times 16 \times 28 \times 28$, with a temporal dimension of 4, 8 and 16, and feature channel dimensions of 1024, 832 and 480, respectively.

\subsection{Hierarchical Network Head}

The features extracted from the I3D model are used as input to a hierarchical network head to predict signs at different temporal resolutions. The hierarchical network head follows a U-Net \cite{ronneberger2015u} design, but instead of increasing the spatial resolution we increase the temporal resolution, making class prediction outputs at different temporal levels by creating skip connections between the different feature layers.

\begin{figure*}
    \centering
    \includegraphics[width=1.0\textwidth]{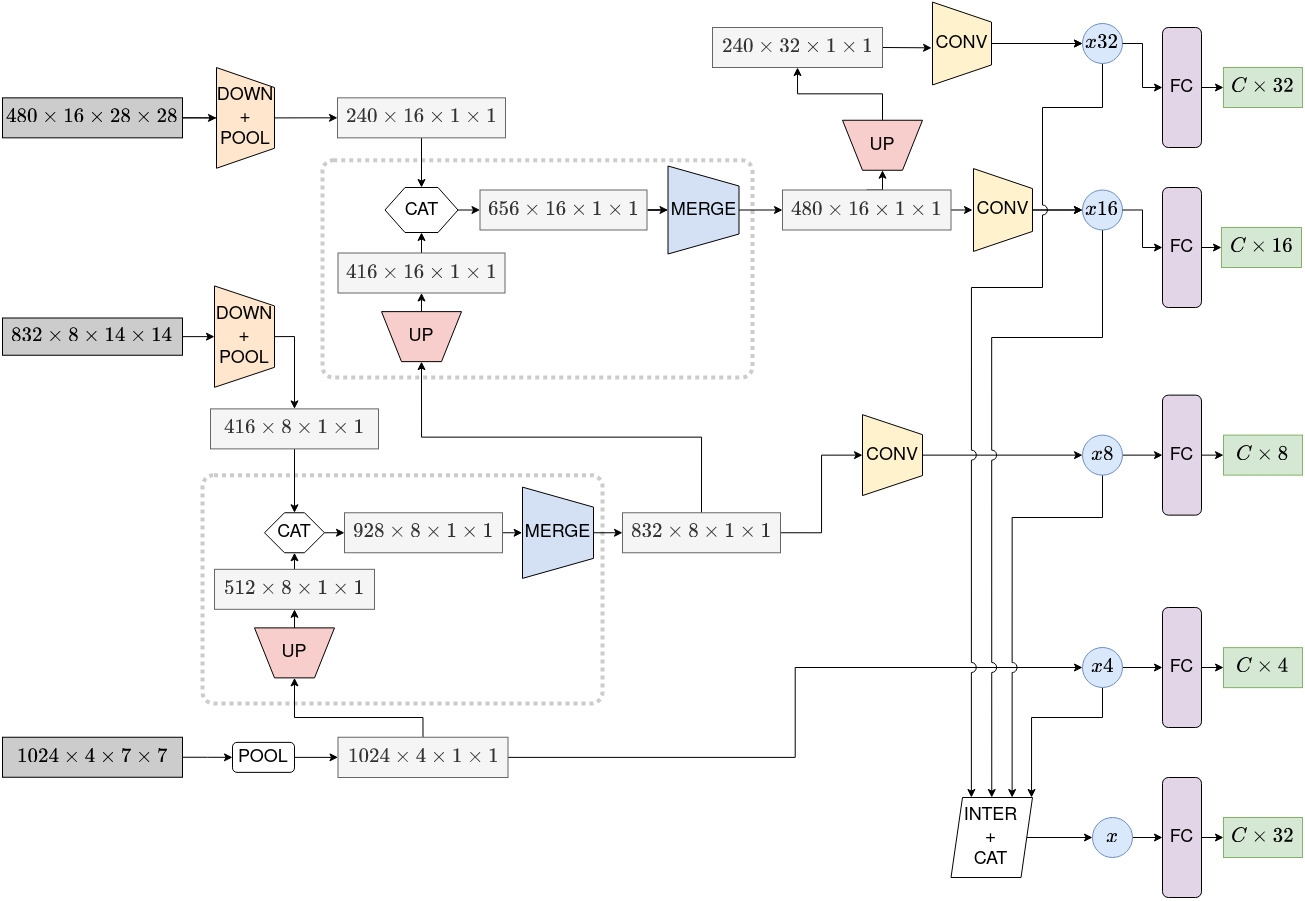}
    \caption{Flow of the introduced hierarchical network head when given a sequence of 32 frames with dimensions $224\times224$. The three inputs are taken from outputs of various stages of the I3D model. The output consists of five temporal segment predictions of various segment lengths}
    \label{fig:hier_arch}
\end{figure*}

An overview describing the flow of the hierarchical network architecture is given in \cref{fig:hier_arch}, where we define the following network blocks as follows:
    
\textbf{POOL} consists of 3D Global Average Pooling which reduces the spatial dimensions to $1\times1$ while keeping the original temporal and feature dimensions.

\textbf{UP} consists of 3D transpose convolution which doubles the input temporal dimension and halves the feature dimension. The temporal dimension is doubled using a kernel size of $(2,1,1)$ and a stride of $(2,1,1)$.

\textbf{CAT} represents concatenating the feature outputs, which is used as a form of skip connection between the different feature layers.

\textbf{MERGE} consists of a 3D convolution layer followed by batch normalisation \cite{ioffe2015batch} and swish activation \cite{ramachandran2017searching} which is repeated twice. The first 3D convolution has a kernel size of $(1,1,1)$ and the same number of output channels as the input channels. The second 3D convolution has a kernel size of $(1,1,1)$ with output channel dimensions being double the input of the lowest feature channel dimension before the respective \textbf{CAT} block. 

\textbf{CONV} consists of a 3D convolution with a kernel size of $(1,1,1)$ keeping the same output dimension as the input dimension.

\textbf{DOWN + POOL} consists of a Residual Block \cite{he2016deep}, replacing the ReLU activation with the swish activation function. The Residual Block halves the feature channel dimension, keeping the output temporal dimensions the same but reducing the spatial dimensions to $7\times7$ by adjusting the spatial stride to $(2,2)$ for the input spatial resolution of $14\times14$ and $(4,4)$ for the input spatial resolution of $28\times28$. This is followed by the \textbf{POOL} block defined above.

\textbf{INTER + CAT} consists of interpolation of the temporal dimension to size 32 using the nearest approach followed by concatenation of the input features.

\textbf{FC} is a fully connected layer with output size of the number of classes ($C$).
\\

As shown in \cref{fig:hier_arch}, the hierarchical network head uses the features extracted from the I3D model as input to coarse-to-fine localisation predictions.
The temporal prediction levels are as follows: \textbf{temporal level 4 ($x4$)} makes 1 prediction every 8 consecutive frames with a total of 4 predictions; \textbf{temporal level 8 ($x8$)} makes 1 prediction every 4 consecutive frames with a total of 8 predictions; \textbf{temporal level 16 ($x16$)} makes 1 prediction every 2 consecutive frames with a total of 16 predictions; \textbf{temporal level 32 ($x32$)} makes 1 prediction every frame with a total of 32 predictions.

An additional \textbf{combined temporal level ($x$)} is created to combine features from all temporal levels, which makes 32 prediction, one prediction for every frame.
This is achieved by temporally interpolating the features from all temporal levels to a temporal dimension of 32 and concatenates the features before a fully connected layer, which produces the interpolated prediction. We therefore obtain a hierarchy of 5 sign localisation predictions at different temporal resolutions.

\subsection{Learning Objectives}
\label{sec:learning_objective}

Cross entropy loss is used as the learning objective, where the target is set to the sign class label, if it exists within the prediction window, else the target is set to an additional \emph{``background''} class. To obtain the final prediction, softmaxed logits from all five levels are temporally interpolated to the original sequence length. We then take their average for each frame to get sign class probabilities. A greedy search is applied which selects the highest probability class as the sign label for each frame. Consecutive frames of the same sign label, excluding the background class, are then selected as the temporal window for the identified sign.

\section{Experiments}

In this section, we evaluate the proposed HS-I3D model on the challenging LSE\_eSaude\_ UVIGO dataset. We first introduce the data set and the sign spotting objective. Then we give an overview of the evaluation metric used to measure the sign spotting performance of the model. Next we demonstrate qualitative results of the proposed approach to give reader more insight. Finally, we perform ablation studies, demonstrating the importance of the hierarchical components of our network and the impact of random data sampling on the precision and recall.

\subsection{LSE\_eSaude\_UVIGO dataset}

LSE\_eSaude\_UVIGO \cite{Enriquez:ECCVW:2022} is a Spanish Sign Language (LSE: Lengua de Signos Española) dataset in the health domain with around 10 hours of continuous sign videos. The dataset contains 10 signers which include seven deaf and three interpreters, captured in studio conditions with a constant blue background. It is partially annotated with the exact location of 100 signs, namely beginning and end timestamps, which were annotated by interpreters and deaf signers.

For our work the experiments are performed using the ChaLearn 2022 MSSL (multiple shot supervised learning) track protocol. MSSL is the classical machine learning track where sign classes are the same in the training, validation and test sets (i.e. no out of vocabulary samples). 

We use the official dataset splits, namely; MSSL\_Train\_Set, which is the training dataset containing around 2.5 hours of videos with annotations of 60 signs performed by five people; MSSL\_Val\_Set, which is the validation dataset containing around 1.5 hours of videos with annotations of the same 60 signs performed by four people; and the MSSL\_Test\_Set, which is the test set with around 1.5 hours of videos with annotations of the same 60 signs performed by four people.

\subsection{Evaluation Metric}

For evaluation we use the matching score per sign instance metric, which is based on the intersection over union (IoU) of the ground-truth and predicted intervals. The IoU results are thresholded using the values that are spread between $0.20$ to $0.80$ with $0.05$ steps (i.e. $\{0.20, 0.25, ..., 0.75, 0.80\}$), yielding a set of 13 true positive, false positives and false negatives scores. These are then summed and used to calculate the final F1-score.

\subsection{Random Sampling Probabilities}

SLR approaches conventionally select consecutive frames around a known sign vocabulary during training. But since the objective is sign spotting with an F1-score evaluation metric, it requires a balance between precision and recall. Some sign videos may have signs outside of vocabulary that are visually similar to the signs in the vocabulary, which will impact the precision if predicted as false positives. We address this issue by introducing random sampling probabilities ($rsp$), where instead of selecting frame regions around only known sign classes, we randomly select consecutive frame regions from other areas in the continuous sign video based on the $rsp$. A higher $rsp$ indicates higher probability of selecting consecutive frames from random locations in the video.

\subsection{Implementation Details}

We use the Inception I3D model \cite{carreira2017quo} as our spatio-temporal backbone, which was pretrained on the large scale BSL1K dataset and then on the WLASL dataset \cite{albanie2020bsl}. The ReLU activation functions are replaced by the Swish activation function \cite{ramachandran2017searching}, as it has been shown to improve results for SLR \cite{jiang2021skeleton}.

During training, HS-I3D is trained in an end-to-end manner. The input to the model is 32 consecutive frames of size $224\times224$ with random data augmentations, such as random cropping, rotation, horizontal flipping, colour jitter and gray scaling. Mixup \cite{zhang2017mixup} is also applied with an $\alpha$ value of $1.0$.
Since the LSE\_eSaude\_UVIGO dataset has large class imbalances, we re-balance the class distributions and under-sample majority classes and over-sample minority classes.

The HS-I3D model is trained for 200 epochs with a batch size of 8 using an Adam optimizer \cite{kingma2014adam} with an initial learning rate of $3\times10^{-4}$ and cosine annealing decay until the model converges. The final predictions of the model are based on the average probability output of the different temporal levels, which are temporally interpolated to a temporal length of 32, to match the frame input size.

Four fold cross validation is used, where each fold is separated by signer ID using the  MSSL\_Train\_Set, always keeping signer number five in the training set due to the large number of annotations for their samples. This training processes is repeated three times with different random sampling probabilities. The $rsp$ values used during training is 0.0, 0.1 and 0.5.

An additional six fold cross validation set is used consisting of a mixture of training and validation dataset and follows the same process as the original four fold cross validation approach.

Therefore an ensemble based on the mean probability outputs of 30 models ($4\times3 + 6\times3$) is used for the final predictions, where we apply a temporal stride of one over the test videos, taking the average probabilities between overlapping time segments.

\subsection{Results}

The results of the HS-I3D model are compared to the baseline approach in \cref{tab:final_results}. Our model was able to surpass the baseline F1-score by 0.307, which is a significant improvement. HS-I3D using the averaged output from all temporal levels has a 0.022 performance improvement over just using the main temporal level $x$ output. 

\begin{table}
    \centering
    \caption{Performance of the ensemble HS-I3D models on the LSE\_eSaude\_UVIGO test set}
    \setlength{\tabcolsep}{5pt}
    \begin{tabular}{l | c}
    
        \toprule
        Model & F1-score \\
        \midrule
        \midrule
        Baseline & 0.300\\
        \midrule
        HS-I3D ([$x$]) & 0.585\\
        HS-I3D (avg[$x$,$x4$,$x8$,$x16$,$x32$]) & \textbf{0.607}\\
        \midrule
    \end{tabular}
    \label{tab:final_results}
\end{table}

\subsection{Ablation Studies}

In this section, we analyze the impact of the hierarchical network head at different levels. Then, we demonstrate the importance and benefits of different random sampling probabilities. We report results with models trained only on the training dataset (MSSL\_Train\_Set) and display results on the validation dataset (MSSL\_Val\_Set).

\subsubsection{Temporal Levels}

In \cref{tab:hier_results}, we demonstrate the impact of the different temporal levels on the precision, recall and overall F1-score. We find that the most coarse temporal level $x4$ which makes a prediction every 8 frames, has the lowest F1-score, while the temporal level $x8$ has the highest F1-score. Averaging the prediction probabilities shows small improvements to the F1-score with the highest precision score compared to the individual temporal levels.

\begin{table}
    \centering
    \caption{Impact of the different temporal levels from the HS-I3D model on the LSE\_eSaude\_UVIGO validation set}
    \setlength{\tabcolsep}{5pt}
    \begin{tabular}{l | c | c | c}
        \toprule
        Temporal Level & Precision & Recall & F1-score \\
        \midrule
        \midrule
        $x4$  & 0.628 & 0.496 & 0.554\\
        $x8$  & 0.631 & \textbf{0.543} & 0.583\\
        $x16$ & 0.643 & 0.531 & 0.581\\
        $x32$ & 0.634 & 0.513 & 0.567\\
        $x$   & 0.636 & 0.498 & 0.558\\
        \midrule
        average ($x4,x8,x16,x32,x$) & \textbf{0.655} & 0.531 & \textbf{0.586}\\
        \midrule
    \end{tabular}
    \label{tab:hier_results}
\end{table}

\begin{figure*}[t]
    \centering
    \includegraphics[width=1.0\textwidth]{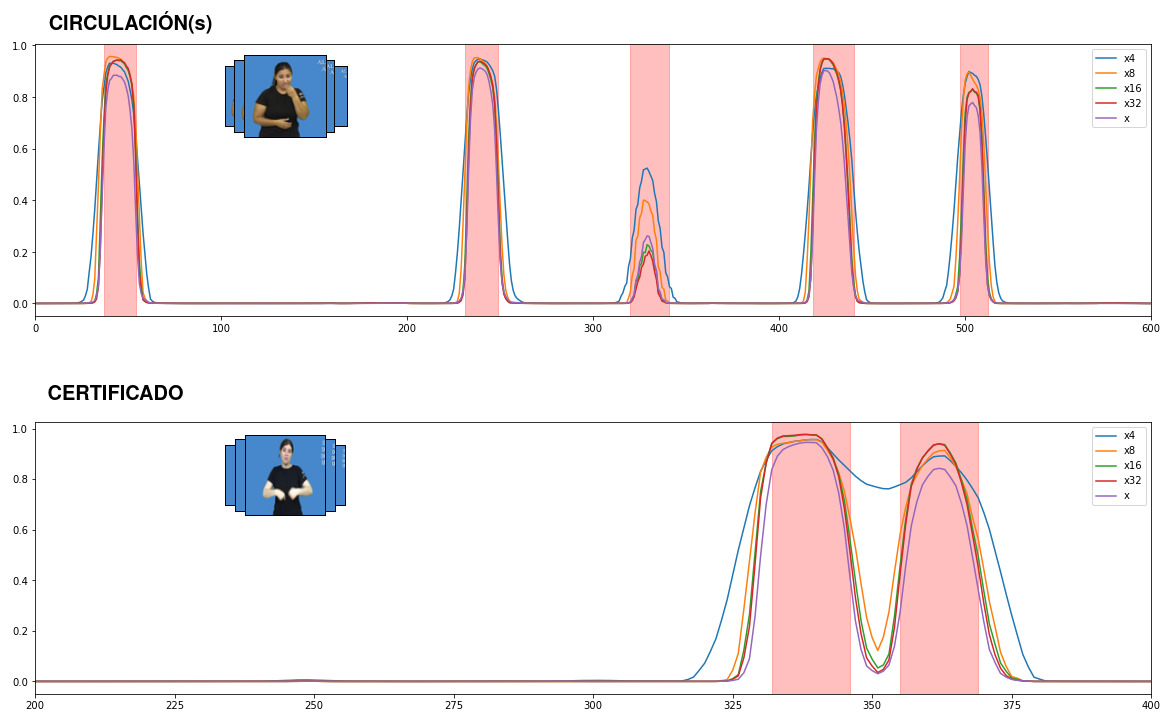}
    \caption{Comparison of prediction probabilities (y-axis) between different temporal levels over the frames (x-axis) in a video sequence. Graphs shows probabilities for the sign CIRCULACIÓN(s) (top) and CERTIFICADO (bottom) over a continuous sign video with the ground truth interval highlighted by the red region.}
    \label{fig:tw_dgrm}
\end{figure*}

In \cref{fig:tw_dgrm} we see the prediction probabilities between the different temporal levels, where in the top diagram we tend to see the temporal level $x4$ probabilities starts to identify the sign sightly earlier and ends slightly later compared to more fine-grained temporal levels. This has a negative impact when the same signs are in close proximity as shown in the bottom figure for the sign CERTIFICADO, where there is not such an easy distinction for temporal level $x4$ compared to other temporal levels.

\subsubsection{Random sampling probability}

In \cref{tab:rsp_results}, we demonstrate the impact of the sampling on the precision, recall and overall F1-score. We note a strong correlation between the precision and random sampling probability, where a $rsp$ of 0.5 has improved precision compared to a model trained with 0.0 $rsp$ with a 0.216 increase in precision. While higher $rsp$ increases the precision, there is a trade off in the recall, where we find a significant decrease of 0.117 in recall. 

\setlength{\tabcolsep}{4pt}
\begin{table}
    \centering
    \caption{Impact of random sampling probabilities on the LSE\_eSaude\_UVIGO validation set}
    \setlength{\tabcolsep}{5pt}
    \begin{tabular}{l | c | c | c}
        \hline\noalign{\smallskip}
        Random Sampling Prob. & Precision & Recall & F1-score \\
        \midrule
        \midrule
        0.0 & 0.439 & \textbf{0.648} & 0.523\\
        0.1 & 0.621 & 0.553 & 0.584\\
        0.5 & \textbf{0.655} & 0.531 & 0.586\\
        \midrule
        Ensemble (0.0,0.1,0.5) & 0.634 & 0.576 & \textbf{0.604}\\
        \midrule
    \end{tabular}
    \label{tab:rsp_results}
\end{table}
\setlength{\tabcolsep}{1.4pt}

\begin{figure*}[t]
    \centering
    \includegraphics[width=1.0\textwidth]{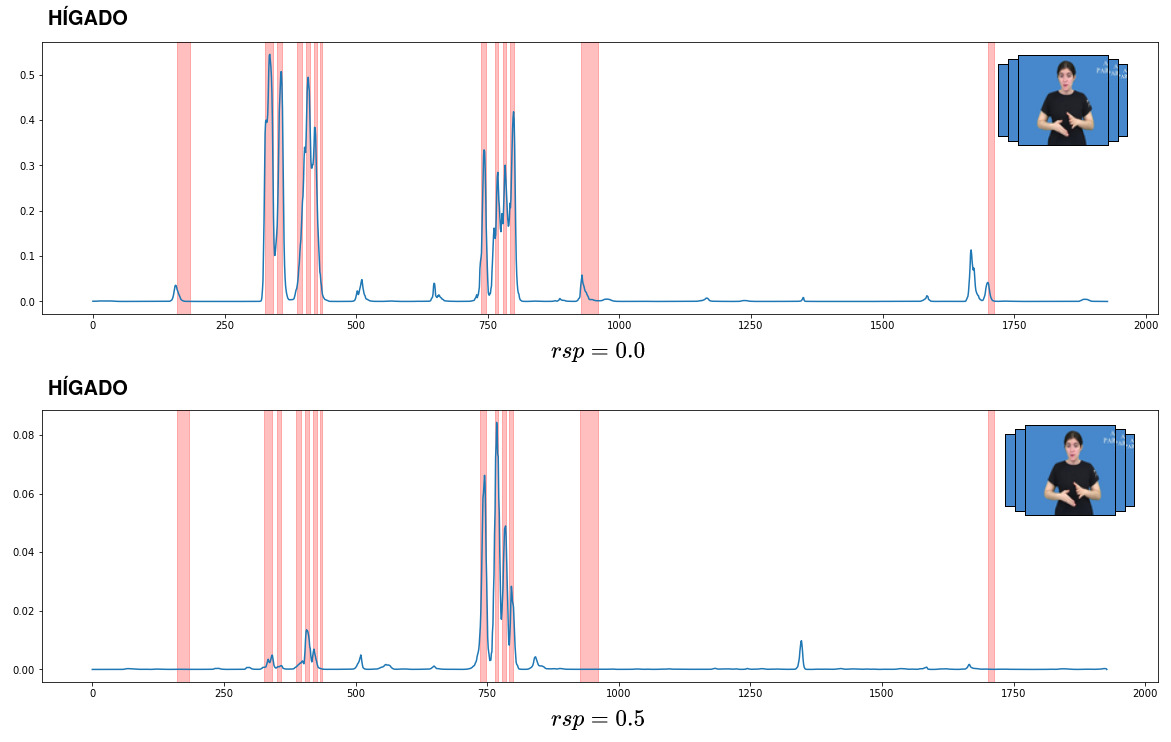}
    \caption{Comparison of probabilities (y-axis) of spotting HÍGADO (blue line) over frames (x-axis) within a continuous sign video with the ground truth interval (red regions). The top graph represents the probabilities obtained with models trained with $rsp=0.0$ while the bottom graph represents the probabilities obtained with models trained with $rsp=0.5$}
    \label{fig:rsp_dgrm}
\end{figure*}

In \cref{fig:rsp_dgrm} we see the prediction probabilities for the sign HÍGADO over time from a continuous sign video. We find that a model trained with an $rsp$ of $0.0$ is able to recall more instances of the sign but at the cost of producing more false positives than a model trained with an $rsp$ of $0.5$.

We find improved results, i.e. better balance between the precision and recall, with an overall higher F1-score, by ensembling 3 different $rsp$ models. The ensembling is done by averaging the prediction probabilities.

\section{Conclusion}

In this paper, we propose the novel Hierarchical Sign I3D (HS-I3D) model, which takes advantage of intermediary layers of an I3D model for coarse-to-fine temporal sign representations. We develop a hierarchical network head, which combines spatio-temporal features for spotting signs at various temporal resolutions.

Our approach is able to make use of existing I3D models pretrained on large scale SLR datasets. We demonstrate the effectiveness of our model and show the importance of random data sampling while noting the trade-off between precision and recall. Our model achieve state-of-the-art performance of $0.607$ F1-score on the ChaLearn 2022 MSSL track for the LSE\_eSaude\_UVIGO dataset winning the Multiple Shot Supervised Learning sign spotting challenge.

Future work involves exploration of better ensemble techniques with the different $rsp$ models and temporal windows instead of the simple averaging prediction probabilities. Additionally, keypoint based models, such as Pose$\rightarrow$Sign or GCNs, can be adapted to work with our hierarchical network head as an alternative input modality for sign spotting.\\

\textbf{Acknowledgments.}
This work received funding from the SNSF Sinergia project 'SMILE II' (CRSII5 193686), the European Union’s Horizon2020 research and innovation programme under grant agreement no. 101016982 'EASIER' and the EPSRC project 'ExTOL' (EP/R03298X/1). This work reflects only the authors view and the Commission is not responsible for any use that may be made of the information it contains.

\clearpage
\bibliographystyle{splncs04}
\bibliography{egbib}
\end{document}